\documentclass[letterpaper,conference]{IEEEtran}
\usepackage{cite}
\usepackage{amsmath,amssymb,amsfonts}
\usepackage{algorithmic}
\usepackage{subfigure}
\usepackage{graphicx}
\usepackage{textcomp}
\usepackage{xcolor}
\usepackage{multirow}
\usepackage{float}
\usepackage{threeparttable}

\def\BibTeX{{\rm B\kern-.05em{\sc i\kern-.025em b}\kern-.08em
    T\kern-.1667em\lower.7ex\hbox{E}\kern-.125emX}}
\begin{document}
\newcommand{\etal}{\textit{et al}.}
\title{Improving Prediction of Need for Mechanical Ventilation using Cross-Attention\\
% \thanks{S.N. is funded by the National Institutes of Health (\#R35GM143121).
}

\author{\IEEEauthorblockN{Anwesh Mohanty}
\IEEEauthorblockA{\textit{Department of Computer Science and Engineering} \\
\textit{University of California San Diego}\\
La Jolla, USA \\
anmohanty@ucsd.edu}
\and
\IEEEauthorblockN{Supreeth P. Shashikumar, Jonathan Y. Lam, Shamim Nemati}
\IEEEauthorblockA{\textit{Division of Biomedical Informatics} \\
\textit{University of California San Diego}\\
La Jolla, USA \\
\{spshashikumar, j7lam, snemati\}@health.ucsd.edu}

}

\maketitle

\begin{abstract}
In the intensive care unit, the capability to predict the need for mechanical ventilation (MV) facilitates more timely interventions to improve patient outcomes. Recent works have demonstrated good performance in this task utilizing machine learning models. This paper explores the novel application of a deep learning model with multi-head attention (FFNN-MHA) to make more accurate MV predictions and reduce false positives by learning personalized contextual information of individual patients. Utilizing the publicly available MIMIC-IV dataset, FFNN-MHA demonstrates an improvement of 0.0379 in AUC and a 17.8\% decrease in false positives compared to baseline models such as feed-forward neural networks. Our results highlight the potential of the FFNN-MHA model as an effective tool for accurate prediction of the need for mechanical ventilation in critical care settings.
\end{abstract}


\section{Introduction}
Mechanical ventilation (MV) is often required when hospitalized patients face respiratory distress or failure and are unable to breathe on their own\cite{ref1,ref2}. Accurate prediction of MV may have an important role in influencing treatment strategies, improving patient outcomes, and optimizing resource utilization\cite{ref3,ref4,ref5}.  Timely initiation of MV\cite{ref6,ref7} can prevent complications and improve patient outcomes, while unnecessary interventions can lead to resource wastage and potential patient discomfort. The inherent complexity of clinical data, marked by dynamic interactions with different patients, presents significant challenges to the development and use of machine learning systems to predict the need for MV.

Recent works in this field have explored deep-learning approaches and traditional machine-learning models for predicting MV. Wang \etal\cite{b4} comprehensively analyzed neural networks and traditional machine learning models for estimating the MV duration in acute respiratory distress syndrome patients. Bendavid \etal\cite{b5} proposed an XGBoost-based model to determine the need to initiate invasive MV in hypoxemic patients. Hsieh \etal\cite{b6} demonstrated that Random Forest models performed better in comparison to artificial neural networks for the prediction of mortality of unplanned extubation patients. 

% In the current literature, traditional methods such as XGBoost\cite{b7} consistently demonstrate superior performance compared to deep learning approaches, despite the widespread adoption and advancements in deep learning methods. This observation raises a critical challenge in the domain, highlighting the need for novel approaches that enable deep learning models to capitalize on their inherent capabilities while addressing the challenges that impede their consistent outperformance. 

Attention-based models have been shown to improve the performance of deep learning models in various domains\cite{b8,attn1,attn2}. The efficacy of attention-based mechanisms is due to their ability to focus on a small subset of the input features relevant to outcome prediction. To this end, our paper introduces the FFNN-MHA model, a feed-forward neural network (FFNN) with a multi-head attention mechanism (MHA) \cite{b8}, designed to navigate the correlations between clinical data. By incorporating multi-head attention mechanisms, the FFNN-MHA model intelligently weighs the relevance of different features, fostering a nuanced understanding of contextual dependencies.

Here, we investigate the addition of attention mechanisms, particularly cross-attention, to enhance the performance of deep learning models for predicting the need for MV. In the following sections, we describe the dataset used in our study, the architecture of the FFNN-MHA model, details regarding model training and evaluation, and a comparative benchmark against various baselines.

% In the following sections, we describe the dataset used in our study, the architecture of the FFNN-MHA model, details regarding model training and evaluation, and a comparative benchmarking against various baselines. Our findings show the effectiveness of attention-based neural networks in accurately predicting respiratory failure within critical care contexts.

 \begin{figure*}[htbp]
  \centering
  \subfigure[Feed Forward neural network]{\includegraphics[scale=0.38]{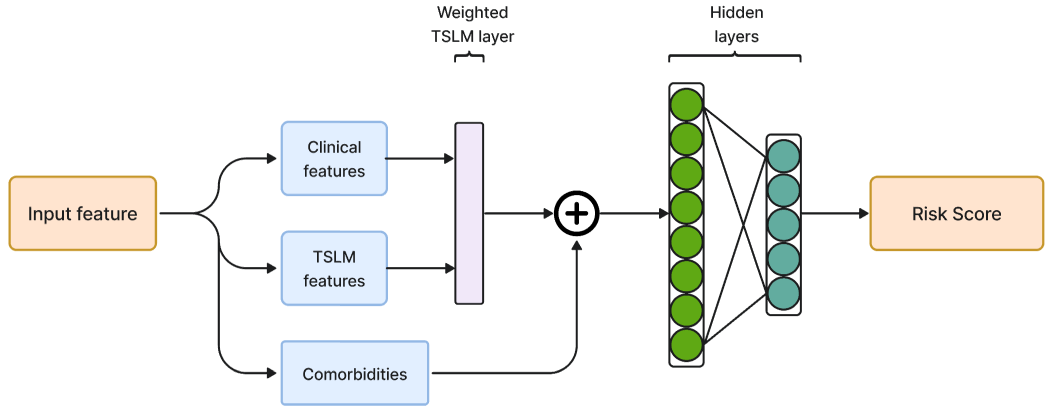}}\quad
  \subfigure[Proposed FFNN-MHA model architecture]{\includegraphics[scale=0.38]{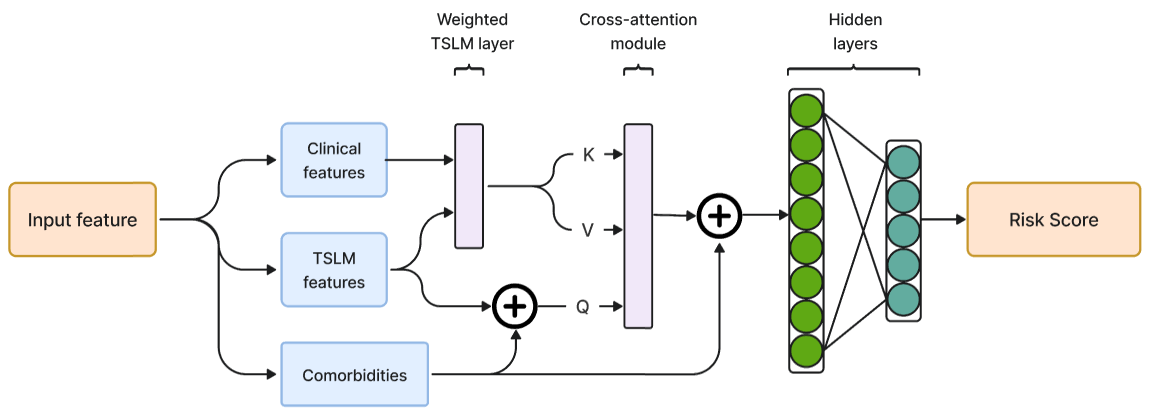}}
  \caption{\textbf{Schematic diagrams for the baseline FFNN and FFNN-MHA models.} (a) Baseline feed-forward neural network with the weighted TSLM layer incorporated from the COMPOSER model. (b) Proposed FFNN-MHA architecture with cross-attention implemented across the clinical features, TSLM features, and comorbidities. The cross-attention module in (b) includes a cross-attention layer followed by a layer normalization applied to the attention output. In both figures, the final output is a value between 0 and 2 indicating the risk score in a patient.}
  \label{arch}
\end{figure*}

\section{Methods}

\subsection{Dataset}

An observational, multicenter cohort consisting of all adult patients of at least 18 years of age admitted to the ICU  was considered in this study from the freely accessible MIMIC-IV dataset \cite{b2}. Patients were excluded if (1) their length of stay was less than 4 h or greater than 20 days, or (2) the start of invasive MV occurred before hour 4 of ICU admission, or (3) if they received noninvasive MV. Institutional review board approval for the data was given by the Beth Israel Deaconess Medical Center (IRB Protocol \#2001P001699) with a waiver of informed consent.

The input features consisted of 8 vital signs measurements (such as heart rate, temperature, etc.), 42 laboratory measurements (such as bicarbonate, pH, calcium, etc.), 6 demographic variables (such as age, gender, etc.), 11 medication categories (such as on-anesthesia, on-anticoagulants, etc.) and 62 comorbidities (such as liver cirrhosis, malignancy, etc.) binned into hourly timestamps. Patients with MV were labeled using a composite score: invasive MV $\leq$24 hours (1 point), and invasive MV $>$24 or $\leq$24 hours with mortality (1 point). For model evaluation, a composite score of $\geq$1 was defined as the positive class. Invasive MV was defined as the first occurrence of simultaneous recording of a fraction of inspired oxygen (FiO2) and positive end-expiratory pressure (PEEP). 

\subsection{FFNN-MHA Model}

The FFNN-MHA model builds upon the architecture of COMPOSER\cite{b1}, incorporating a novel approach to leverage the Time Since Last Measured (TSLM) features in a more refined manner, as shown in Figure \ref{arch}. While COMPOSER was initially designed for sepsis prediction, our focus shifted to utilizing FFNN-MHA for predicting the need for MV in patients. In COMPOSER, the TSLM layer consists of a weighted input layer designed to scale the latest measured value of a clinical variable based on the duration since its last measurement. This scaling is controlled by a parameter learned from the data, to appropriately account for the age of an imputed feature while preventing the model from directly exploiting the frequency of measurements.

% Attention-based models have been shown to improve the performance of deep learning models in various domains\cite{b8,attn1,attn2}. The efficacy of attention-based mechanisms is because they provide deep learning models with the ability to focus on a small subset of the input features relevant to outcome prediction. 
In this work, we utilize the popular multi-head attention mechanism to apply attention across the input clinical variables at any given time. In the FFNN-MHA model, we use the extracted TSLM features from the weighted TSLM layer as queries to the attention module. We further augment the queries with the comorbidities of each patient to allow the FFNN-MHA model to capture contextual dependencies between clinical features, TSLM features, and patient comorbidities. By integrating this nuanced relationship into the attention mechanism, the FFNN-MHA model goes beyond the conventional use of the TSLM layer, enhancing its ability to discern temporal and personalized patient patterns while mitigating the risk of overfitting institutional-specific workflow practices and care protocols.

\subsection{Model development and training}
In our evaluation of attention-based training strategies, we explored various combinations of inputs to the attention module to achieve optimal prediction performance. In particular, we assessed self-attention (SA) and cross-attention (CA) mechanisms. For the FFNN-MHA model, we used the TSLM features along with patient comorbidities as the query vector with the input clinical data serving as key and value. 

All of the FFNN models (FFNN, FFNN+SA, FFNN+CA, FFNN-MHA) used in this study consisted of a three-layered feedforward neural network (of size 100, 80, and 60) trained to predict the onset of MV up to 24 hours in advance. For the FFNN-MHA model, we set the key dimension to 150 and used a total of 3 heads for the attention module. The final output of the FFNN-MHA model for each patient is a risk score, a numerical prediction between 0 and 2, where a risk score close to 0 indicates a healthy patient and a score close to 2 indicates a high necessity for MV in the patient.

The parameters of the FFNN models were randomly initialized and trained on the training data with L1-L2 regularization and dropout to avoid overfitting. The FFNN-MHA model was trained with RMSE loss for 300 epochs using Adam optimizer\cite{adam} with a batch size of 3000 and a learning rate of 0.006. The model with the best performance, measured by Area Under the Receiver Operating Character Curve (AUC) on the validation dataset, was selected. All of the hyperparameters were optimized using Bayesian hyperparameter optimization. The entire cohort was randomly split into training (80\%) and testing (20\%) cohorts.

% This approach demonstrated a nuanced understanding of contextual dependencies and information sharing between the TSLM queries and the clinical data, enhancing the FFNN-MHA model's overall performance in predicting MV requirements.

% \subsection{Model Inference}

% In the inference phase, the trained FFNN-MHA model processes new data from the test split of the MIMIC dataset to predict the likelihood of MV. The model utilizes the multi-headed attention module and feedforward neural network to extract and analyze relevant patterns, providing valuable insights into patient outcomes. During the prediction task, the FFNN-MHA model outputs a probability score indicating confidence in the prediction of MV requirements. This probabilistic output offers healthcare providers valuable insights into patient outcomes, assisting in the identification of individuals at risk. The model's adaptability to real-time data and its sophisticated integration of temporal context through the TSLM layer and attention mechanism contribute to its effectiveness in critical care scenarios. The FFNN-MHA model stands as a robust tool for healthcare practitioners, providing actionable predictions to guide timely interventions and optimize patient care. 

\subsection{Evaluation metrics}

For all continuous variables, we have reported the median and interquartile ranges. For binary variables, we have reported percentages. The AUC, Area Under the Precision-Recall Curve (AUCpr), Specificity (SPC), Positive Predictive Value (PPV), and number of False positive (FP) alarms at 80\% Sensitivity level were used to measure model performance. All of the above metrics were measured at the 1-hour window level. The AUC was calculated under an end-user clinical response policy in which the model was silenced for 6 hours after an alarm was fired. The significance between the AUCs was determined using DeLong's test\cite{delong}.

% FFNN-MHA's performance is comprehensively assessed using five distinct metrics: Area Under the Receiver Operating Characteristic Curve (AUC), Specificity (SPC) at 80\% sensitivity level, Positive Predictive Value (PPV), False Positives, and Area Under the Precision-Recall Curve (AUCpr). 

% This multi-faceted evaluation approach ensures a thorough examination of the model's predictive capabilities. AUC highlights the model's effectiveness in discriminating between different outcomes. Specificity and PPV provide insights into the model's ability to minimize false positives and false negatives, crucial considerations in clinical applications. Additionally, AUCpr offers a nuanced perspective on the model's performance, especially in scenarios with imbalanced class distribution. The collective assessment across these metrics reinforces the robustness and reliability of FFNN-MHA in the context of predicting MV requirements.

\section{Results and Discussions}

\subsection{Patient characteristics}

After applying the exclusion criteria, a total of 54,636 ICU patients were included in the study of which 80.74\% were non-ventilated and 19.26\% required ventilation. The median [interquartile] length of stay in the ICU for patients on MV was higher compared to non-ventilated patients, 92 [49 - 173.8] hours vs 42.6 [25 - 74.7] hours. The in-patient mortality rate was 15.74\%  for ventilated patients and 8.94\% for non-ventilated patients. Table \ref{tab2} summarizes the patient characteristics of the cohort used in our study.

\begin{table}[]
\caption{Patient characteristics of the study cohort}
\label{results}
\begin{center}
\begin{tabular}{|c|cc|}
\hline
Characteristic & \multicolumn{1}{c|}{Nonventilated} & Ventilated \\ \hline
Patients & \multicolumn{1}{c|}{44,112 (80.74\%)} & 10,524 (19.26\%) \\
Age, (years) & \multicolumn{1}{c|}{64 (52-76)} & 65 (54-75) \\
Male sex & \multicolumn{1}{c|}{24,013 (54.44\%)} & 6,513 (61.89\%) \\ \hline
Race & \multicolumn{2}{c|}{} \\ \hline
White & \multicolumn{1}{c|}{29,985 (67.97\%)} & 6,950 (66.04\%) \\
Hispanic & \multicolumn{1}{c|}{1,761 (3.99\%)} & 384 (3.65\%) \\
Black & \multicolumn{1}{c|}{5,140 (11.65\%)} & 916 (8.70\%)\\
Asian & \multicolumn{1}{c|}{1,359 (3.08\%)} & 281 (2.67\%)\\
Native American & \multicolumn{1}{c|}{81 (0.18\%)} & 19 (0.18\%)\\
Unknown/Declined to answer & \multicolumn{1}{c|}{4,063 (9.21\%)} & 1,544 (14.67\%)\\
Other & \multicolumn{1}{c|}{1,723 (3.91\%)} & 430 (4.09\%)\\ \hline
ICU LOS, (hours) & \multicolumn{1}{c|}{42.6 (25-74.7)} & 92 (49-173.8) \\
CCI & \multicolumn{1}{c|}{4 (2-7)} & 4 (3-6) \\
SOFA & \multicolumn{1}{c|}{2 (1-4)} & 3 (2-4) \\
Inpatient mortality & \multicolumn{1}{c|}{3,944 (8.94\%)} & 1,656 (15.74\%)\\
\begin{tabular}[c]{@{}c@{}}Time from ICU admission\\  to start of ventilation, (hours)\end{tabular} & \multicolumn{1}{c|}{N/A} & 16 (8-41) \\ \hline
\end{tabular}
\label{tab2}
\end{center}
\end{table}

\subsection{Performance Evaluation}

\begin{figure}[htbp]
    \centering
    \includegraphics[width=7cm]{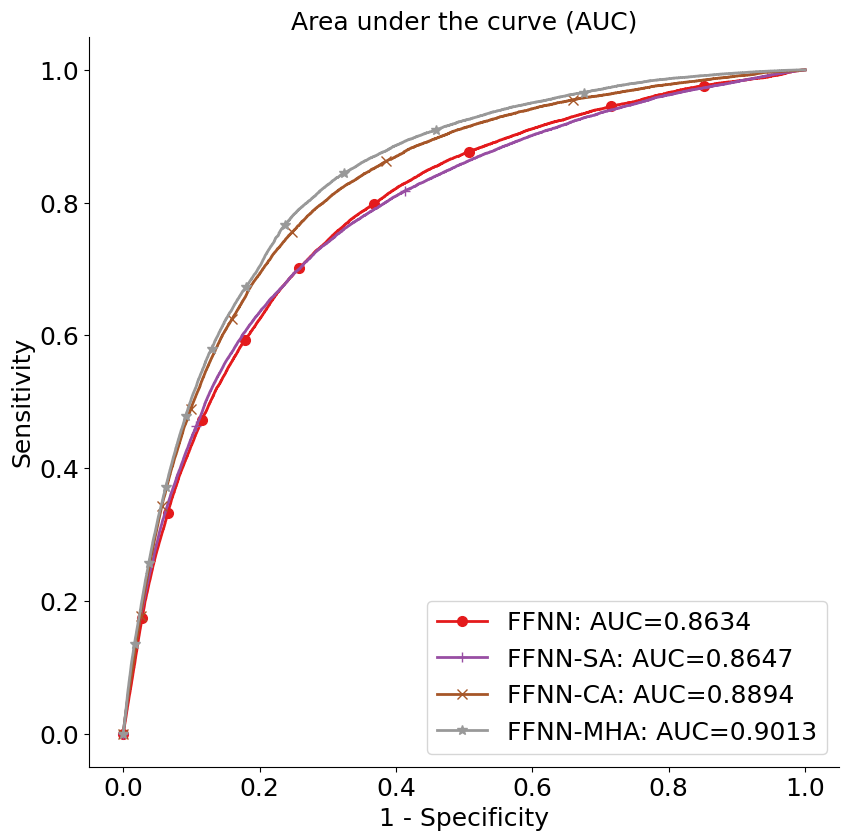}
    \caption{\textbf{AUC plots for FFNN variations considered in this study.} }
    % Self-attention (SA) doesn't give any visible boosts to performance but cross-attention (CA) gives significantly better results compared to the base FFNN. FFNN-MHA gives the best performance.}
    \label{aucplot}
\end{figure}

The baseline FFNN model achieved an AUC of 0.8634 on the testing set (AUC of 0.8794 on the training set) with the specificity (SPC) and positive predictive value (PPV) of 76.51\% and 9.8\% respectively (Table \ref{results}). The feed-forward neural network with self-attention achieved a testing set AUC of 0.8647 (AUC of 0.8801 on the training set). Including a cross-attention module as opposed to a self-attention module resulted in a substantial performance improvement (testing set AUC of 0.8894 vs 0.8647). The TSLM features were used as query vectors and clinical features were used as key vectors for the cross attention module. The final FFNN-MHA model consisted of a cross-attention module with TSLM features and comorbidities used as query vectors and clinical features used as key vectors. We observed that the FFNN-MHA model achieved the highest performance in comparison to all the models with an AUC of 0.9013 (AUC of 0.9312 on the training set), SPC, and PPV of 85.10\% and 12.04\% respectively. AUC plots for all the models are shown in Figure \ref{aucplot}, highlighting the outperformance of FFNN-MHA compared to other models.

% Table \ref{tab1} reports the AUC, SPC, PPV and \#FP for all the models. 

% For the comprehensive benchmarking of predictive performance, we compare our model against other regression models on time-series datasets like logistic regression and XGBoost. We also include performances on standard feed-forward neural networks (FFNN) along with their variations including models with self-attention and cross-attention. Notably, incorporating cross-attention into the FFNN model resulted in a substantial boost in performance. The cross-attention was applied with the TSLM features as queries and clinical features as keys. This compelling improvement served as a pivotal motivator to delve further into the application of cross-attention across different input features in the development of the FFNN-MHA model, with the best results being obtained when a combination of TSLM features along with patient comorbidities was used as queries for the attention module. 

\begin{table}[htbp]
% \begin{threeparttable}[b]
\caption{Comparison of model performance. 
}
\label{results}
\begin{center}
\begin{tabular}{|c|c|c|c|c|}
\hline
\textbf{Model} & \textbf{AUC} & \textbf{SPC (\%)} & \textbf{PPV (\%)} & \textbf{\#FP}\\
\hline
%Ridge Regression & 0.8562 &  76.51 & 9.87 & 49642 \\
%XGBoost Regressor & 0.8934 &  83.80 & 11.61 & 41504\\
FFNN & 0.8634 &  78.02 & 10.15 & 48233\\
FFNN + SA & 0.8647 &  77.95 & 10.13 & 48301\\
FFNN + CA & 0.8894 &  83.77 & 11.55 & 41479\\
FFNN-MHA & \textbf{0.9013} & \textbf{85.10} & \textbf{12.04} & \textbf{39639}\\
\hline
\end{tabular}
\begin{tablenotes}
    \item [1] FFNN: Feedforward neural network, FFNN+SA: FFNN with self-attention, FFNN+CA: FFNN with cross attention, FFNN-MHA: Proposed model
    \item [2] AUC: Area Under the Curve, SPC: Specificity, PPV: Positive predictive value, \#FP: Number of False positives.
    \item [3] SPC, PPV and \#FP was measured at 80\% Sensitivity
\end{tablenotes}
\label{tab1}
\end{center}
% \end{threeparttable}
\end{table}

% \begin{table}[htbp]
% \caption{Comparison of model performances}
% \label{results}
% \begin{center}
% \begin{tabular}{|c|c|c|c|c|c|}
% \hline
% \textbf{Model} & \textbf{AUC} & \textbf{AUCpr} & \textbf{SPC} & \textbf{PPV} & \textbf{FP}\\
% \hline
% Linear Regression & 0.8562 & 0.1826 & 0.7651 & 0.0987 & 49642 \\
% XGBoost Regressor & 0.8934 & 0.2350 & 0.8380 & 0.1161 & 41504\\
% FFNN & 0.8634 & 0.1746 & 0.7802 & 0.1015 & 48233\\
% FFNN + SA & 0.8647 & 0.1721 & 0.7795 & 0.1013 & 48301\\
% FFNN + CA & 0.8894 & 0.1925 & 0.8377 & 0.1155 & 41479\\
% FFNN-MHA & \textbf{0.9013} & \textbf{0.2068} & \textbf{0.8510} & \textbf{0.1204} & \textbf{39639}\\
% \hline
% \end{tabular}
% \label{tab1}
% \end{center}
% \end{table}

The AUC from the FFNN-MHA model was significantly higher than the FFNN model (0.9013 vs 0.8647, $p < 0.0001$). The FFNN-MHA model demonstrated a remarkable 17.8\% reduction in the number of false positives in comparison to the baseline FFNN model (39,639 FPs vs 48,233). Utilizing a cross-attention module resulted in a decrease in false positives in comparison to using a self-attention module (41,479 FPs vs 48,301 FPs).

% In the evaluation of model performance on the test cohort of the MIMIC dataset, the FFNN-MHA model showcased significant advancements over the baseline models. Specifically, in comparison to the standard FFNN model, the FFNN-MHA model achieved a noteworthy 3.79\% improvement in AUC, signaling enhanced discriminative power. Furthermore, there was a substantial 7.08\% boost in specificity (SPC), indicating an improved ability to correctly identify true negatives. Most impressively, the FFNN-MHA model demonstrated a remarkable 17\% reduction in false positives, highlighting its effectiveness in minimizing erroneous predictions. 

% Apart from these, FFNN-MHA outperformed other traditional machine learning techniques like ridge regression and XGBoost by a significant margin. %, implying that such models were not able to leverage the information present in the dataset well. 

\subsection{Interpretability analysis}

\begin{figure}[htbp]
    \centering
    \includegraphics[width=7cm]{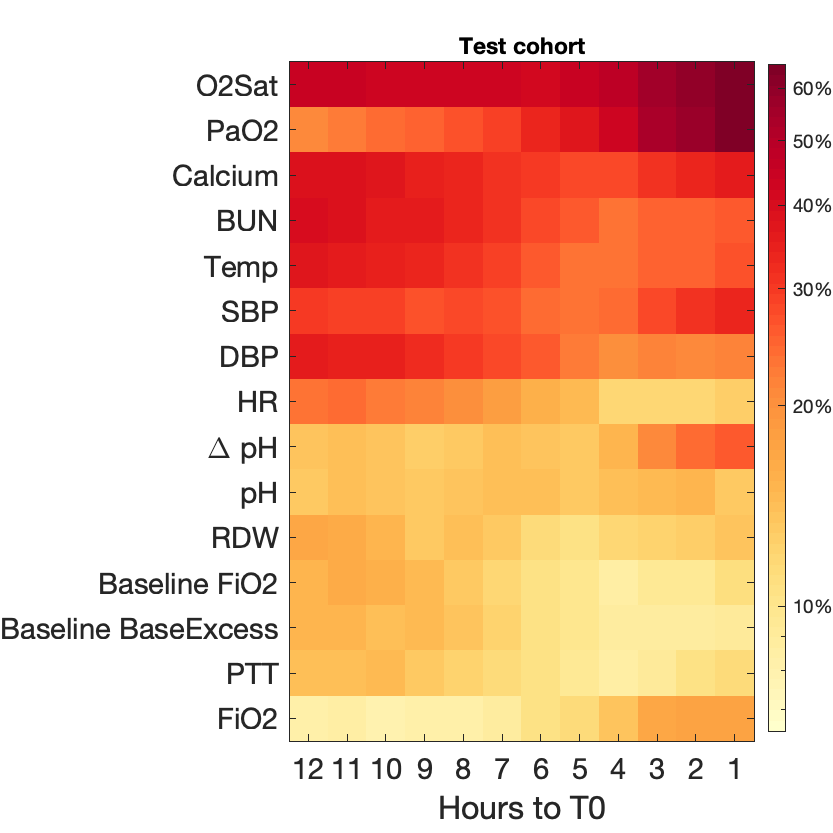}
    \caption{\textbf{Heatmap showing population level plot of contributing factors to the increase in model risk score.} For example, $O_2Sat$ was identified as top contributing factor in $\sim$50\% of ventilated patients 12 hours prior to $T_0$ while it was a top contributing factor in $\sim$60\% of ventilated patients 1 hour prior to $T_0$. The x-axis represents hours before the onset time of MV. The y-axis represents the top factors (sorted by the magnitude of relevance score) across the patient populations.}
    \label{heatmap}
\end{figure}

% We interpret the model by calculating the relevance score of each input variable for every predicted risk score. Figure.\ref{heatmap} shows a heatmap of the top 15 variables contributing to the increase in risk score up to 12 hours before intubation for the testing cohort. As expected, factors like $O_2Sat$, $PaO_2$ and Calcium, among others, contribute highly to the risk score. We have only shown dynamically changing variables in Fig.\ref{heatmap}. Among static variables, sex and duration spent in the hospital were among the top factors. The Transposer model successfully leverages this information to make a more accurate prediction on the test dataset. 

We facilitated model interpretation by computing relevance scores\cite{b1} for each input variable with respect to the predicted risk score. In Figure \ref{heatmap}, a heatmap is presented, highlighting the top 15 variables contributing to the escalation of the risk score up to 12 hours before intubation in the testing cohort. It can be seen that clinical variables such as $O_2Sat$, $PaO_2$, and $Calcium$\cite{calc} prominently contribute to the increase in risk score. The heatmap specifically showcases the fact that the contribution of clinical variables toward risk score can vary temporally in the hours leading up to the time of MV.
% The FFNN-MHA model adeptly incorporates this information, leading to more accurate predictions on the test dataset. 

\section{Conclusion}

% \begin{figure}[htbp]
%     \centering
%     \includegraphics[width=7cm]{shap_values_summary.png}
%     \caption{SHAP values of features that affect the output the most}
%     \label{shap}
% \end{figure}

% The utilization of SHAP (SHapley Additive exPlanations) values\cite{b3} in our analysis have unveiled key factors influencing the predictions of the Transposer model as shown in Fig.\ref{shap}. This analysis provides a detailed understanding of the variables shaping the model's output, offering crucial insights into the complex relationships within the clinical data. The integration of attention mechanisms, such as self and cross-attention, emerges as a pivotal strategy to harness and prioritize the features highlighted by the SHAP analysis. By aligning attention with the most influential factors, the Transposer model achieves enhanced predictive accuracy, leveraging attention to discern intricate patterns and dependencies.

% The synergy between attention mechanisms and SHAP analysis is particularly pronounced in the Transposer model's performance. The inclusion of cross attention, utilizing the output from the weighted TSLM layer in the Composer model, stands out as a key innovation. This model configuration, with a key dimension of 150 and four attention heads, not only aligns with SHAP-identified influential factors but also showcases superior performance. The intelligent allocation of attention to critical features enhances the model's ability to predict MV requirements, reflecting a harmonious interplay between attention mechanisms and interoperability.

In this study, we demonstrated that a feedforward neural network (FFNN) with a multi-head cross-attention module achieved significantly higher performance for the prediction of the need for MV in comparison to a baseline FFNN. We observed that utilizing comorbidity features in addition to TSLM features for query vectors improved model performance. Thus, the final FFNN-MHA model consisted of the combined TSLM features and comorbidity features as query vectors, and the clinical features as key vectors. The utilization of multi-head attention allowed the model to efficiently extract temporal and patient-specific information by understanding the contextual dependencies within the clinical data.
% The FFNN-MHA demonstrated a significant improvement in the number of false positives in comparison with a baseline FFNN (39,639 vs 48,233).

% deptly extract relevant information from diverse streams of inputs. This approach, surpassing the capabilities of a simple FFNN, contributed to the FFNN-MHA model's remarkable accuracy in predicting MV requirements within the clinical context. The effectiveness of multi-headed attention in integrating information from different sources reinforces its pivotal role in enhancing the accuracy and reliability of predictive modeling in complex healthcare scenarios.

% The addition of cross-attention to the standard FFNN model yields notable improvements in predictive accuracy as seen in Table \ref{results}. However, the most significant enhancement is observed when employing cross-attention, leveraging both clinical features and the TSLM features in the model. This cross-attention mechanism, a fundamental component of the FFNN-MHA model, has demonstrated the most substantial improvement in predictive performance. The utilization of multi-headed attention allows the model to adeptly extract relevant information from diverse streams of inputs. This approach, surpassing the capabilities of a simple FFNN, contributes to the FFNN-MHA model's remarkable accuracy in predicting MV requirements within the clinical context. The effectiveness of multi-headed attention in integrating information from different sources reinforces its pivotal role in enhancing the accuracy and reliability of predictive modeling in complex healthcare scenarios.

The inclusion of comorbidity features to improve the performance of the FFNN-MHA model strongly suggests that patient comorbidity is a pivotal feature to incorporate in the cross-attention mechanism, emphasizing its importance in refining contextual dependencies and enhancing the model's predictive capabilities. The FFNN-MHA model, by leveraging patient comorbidities alongside TSLM features, showcases its capacity to capture individualized risk factors and demonstrates the significance of attention in predicting the need for MV.

While the FFNN-MHA model demonstrated good results in predicting the need for MV in the MIMIC-IV cohort, its performance across other cohorts has not been validated. An additional limitation is the possibility of mislabeling MV using the simultaneous recording of FiO2 and PEEP as MV has to be inferred from these measurements in the MIMIC-IV dataset. Future work includes external validation of FFNN-MHA on other MV datasets and assessing how well the model architecture performs on other clinical prediction tasks.

% The addition of patient comorbidities in the query for cross-attention yielded a notable improvement, showcasing the model's adaptability to personalized information. As per Table.\ref{results}, the performance comparison between Composer with cross attention and Transposer implies that the primary distinguishing factor is the inclusion of patient comorbidity in the latter. This result strongly suggests that patient comorbidity is a pivotal feature to incorporate in the cross-attention mechanism, emphasizing its importance in refining contextual dependencies and enhancing predictive capabilities. The Transposer model, by leveraging patient comorbidities alongside TSLM features, showcases its capacity to capture individualized risk factors and demonstrates the significance of personalization in predicting MV requirements.

\section*{Acknowledgment}

S.N. is funded by the National Institutes of Health (\#R01LM013998, \#R01HL157985, \#R35GM143121). A.M, S.P.S and J.Y.L have no sources of funding to declare. S.N and S.P.S are co-founders of a UCSD start-up, Healcisio Inc., which is focused on the commercialization of advanced analytical decision support tools. The opinions or assertions contained herein are the private ones of the author and are not to be construed as official or reflecting the views of the NIH or any other agency of the US Government.

\bibliography{main}

\begin{thebibliography}{10}
\providecommand{\url}[1]{#1}
\csname url@samestyle\endcsname
\providecommand{\newblock}{\relax}
\providecommand{\bibinfo}[2]{#2}
\providecommand{\BIBentrySTDinterwordspacing}{\spaceskip=0pt\relax}
\providecommand{\BIBentryALTinterwordstretchfactor}{4}
\providecommand{\BIBentryALTinterwordspacing}{\spaceskip=\fontdimen2\font plus
\BIBentryALTinterwordstretchfactor\fontdimen3\font minus \fontdimen4\font\relax}
\providecommand{\BIBforeignlanguage}[2]{{%
\expandafter\ifx\csname l@#1\endcsname\relax
\typeout{** WARNING: IEEEtran.bst: No hyphenation pattern has been}%
\typeout{** loaded for the language `#1'. Using the pattern for}%
\typeout{** the default language instead.}%
\else
\language=\csname l@#1\endcsname
\fi
#2}}
\providecommand{\BIBdecl}{\relax}
\BIBdecl

\bibitem{ref1}
C.~Summers, R.~Todd, G.~Vercruysse, and F.~Moore, ``Acute respiratory failure,'' pp. 576--586, 01 2022.

\bibitem{ref2}
C.~Roussos and A.~Koutsoukou, ``Respiratory failure,'' \emph{European Respiratory Journal}, vol.~22, no. 47 suppl, pp. 3s--14s, 2003.

\bibitem{ref3}
R.~Inglis, E.~Ayebale, and M.~Schultz, ``Optimizing respiratory management in resource-limited settings,'' \emph{Current Opinion in Critical Care}, vol.~25, p.~1, 12 2018.

\bibitem{ref4}
\BIBentryALTinterwordspacing
S.~R. Wilcox, J.~B. Richards, A.~Genthon, M.~S. Saia, H.~Waden, J.~D. Gates, M.~N. Cocchi, S.~J. McGahn, M.~Frakes, and S.~K. Wedel, ``Mortality and resource utilization after critical care transport of patients with hypoxemic respiratory failure,'' \emph{Journal of Intensive Care Medicine}, vol.~33, no.~3, pp. 182--188, 2018. [Online]. Available: \url{https://doi.org/10.1177/0885066615623202}
\BIBentrySTDinterwordspacing

\bibitem{ref5}
C.~Williamson, J.~Hadaya, A.~Mandelbaum, A.~Verma, M.~Gandjian, R.~Rahimtoola, and P.~Benharash, ``Outcomes and resource use associated with acute respiratory failure in safety net hospitals across the united states,'' \emph{Chest}, vol. 160, 02 2021.

\bibitem{ref6}
\BIBentryALTinterwordspacing
S.~van Diepen, J.~S. Hochman, A.~Stebbins, C.~L. Alviar, J.~H. Alexander, and R.~D. Lopes, ``{Association Between Delays in Mechanical Ventilation Initiation and Mortality in Patients With Refractory Cardiogenic Shock},'' \emph{JAMA Cardiology}, vol.~5, no.~8, pp. 965--967, 08 2020. [Online]. Available: \url{https://doi.org/10.1001/jamacardio.2020.1274}
\BIBentrySTDinterwordspacing

\bibitem{ref7}
\BIBentryALTinterwordspacing
R.~E. Freundlich, G.~Li, A.~Leis, and M.~Engoren, ``{Factors Associated With Initiation of Mechanical Ventilation in Patients With Sepsis: Retrospective Observational Study},'' \emph{American Journal of Critical Care}, vol.~32, no.~5, pp. 358--367, 09 2023. [Online]. Available: \url{https://doi.org/10.4037/ajcc2023299}
\BIBentrySTDinterwordspacing

\bibitem{b4}
Z.~Wang, L.~Zhang, T.~Huang, R.~Yang, H.~Cheng, H.~Wang, H.~Yin, and J.~Lyu, ``Developing an explainable machine learning model to predict the mechanical ventilation duration of patients with ards in intensive care units,'' \emph{Heart \& Lung}, vol.~58, pp. 74--81, 03 2023.

\bibitem{b5}
I.~Bendavid, L.~Statlender, L.~Shvartser, S.~Teppler, R.~Azullay, R.~Sapir, and P.~Singer, ``A novel machine learning model to predict respiratory failure and invasive mechanical ventilation in critically ill patients suffering from covid-19,'' \emph{Scientific Reports}, vol.~12, 06 2022.

\bibitem{b6}
M.~Hsieh, M.~Hsieh, C.-M. Chen, C.-C. Hsieh, C.-M. Chao, and C.-C. Lai, ``Comparison of machine learning models for the prediction of mortality of patients with unplanned extubation in intensive care units,'' \emph{Scientific Reports}, vol.~8, 11 2018.

\bibitem{b8}
A.~Vaswani, N.~Shazeer, N.~Parmar, J.~Uszkoreit, L.~Jones, A.~N. Gomez, L.~Kaiser, and I.~Polosukhin, ``Attention is all you need,'' 2023.

\bibitem{attn1}
\BIBentryALTinterwordspacing
Y.~Cui, Z.~Chen, S.~Wei, S.~Wang, T.~Liu, and G.~Hu, ``Attention-over-attention neural networks for reading comprehension,'' in \emph{Proceedings of the 55th Annual Meeting of the Association for Computational Linguistics (Volume 1: Long Papers)}.\hskip 1em plus 0.5em minus 0.4em\relax Association for Computational Linguistics, 2017. [Online]. Available: \url{http://dx.doi.org/10.18653/v1/P17-1055}
\BIBentrySTDinterwordspacing

\bibitem{attn2}
J.~Chorowski, D.~Bahdanau, D.~Serdyuk, K.~Cho, and Y.~Bengio, ``Attention-based models for speech recognition,'' 2015.

\bibitem{b2}
A.~Johnson, L.~Bulgarelli, L.~Shen, A.~Gayles, A.~Shammout, S.~Horng, T.~Pollard, S.~Hao, B.~Moody, B.~Gow, L.-w. Lehman, L.~Celi, and R.~Mark, ``Mimic-iv, a freely accessible electronic health record dataset,'' \emph{Scientific Data}, vol.~10, p.~1, 01 2023.

\bibitem{b1}
S.~Shashikumar, G.~Wardi, A.~Malhotra, and S.~Nemati, ``Artificial intelligence sepsis prediction algorithm learns to say “i don’t know”,'' \emph{npj Digital Medicine}, vol.~4, 12 2021.

\bibitem{adam}
D.~Kingma and J.~Ba, ``Adam: A method for stochastic optimization,'' \emph{International Conference on Learning Representations}, 12 2014.

\bibitem{delong}
\BIBentryALTinterwordspacing
E.~R. DeLong, D.~M. DeLong, and D.~L. Clarke-Pearson, ``Comparing the areas under two or more correlated receiver operating characteristic curves: A nonparametric approach,'' \emph{Biometrics}, vol.~44, no.~3, pp. 837--845, 1988. [Online]. Available: \url{http://www.jstor.org/stable/2531595}
\BIBentrySTDinterwordspacing

\bibitem{calc}
C.~Thongprayoon, W.~Cheungpasitporn, A.~Chewcharat, M.~Mao, and K.~Kashani, ``Serum ionised calcium and the risk of acute respiratory failure in hospitalised patients: a single-centre cohort study in the usa,'' \emph{BMJ Open}, vol.~10, p. e034325, 03 2020.

\end{thebibliography}

\end{document}